\documentclass[lettersize,journal]{IEEEtran}
\usepackage{amsmath,amsfonts}
\usepackage{array}
\usepackage[caption=false,font=normalsize,labelfont=sf,textfont=sf]{subfig}
\usepackage{textcomp}
\usepackage{stfloats}
\usepackage{url}
\usepackage{verbatim}
\usepackage{graphicx}
\usepackage{algorithm}
\usepackage{algorithmicx} 
\usepackage{algpseudocode} 
\usepackage{setspace}
\usepackage{cite}
\usepackage{color}
\usepackage{booktabs}
\usepackage{array}
\usepackage{multirow}
\usepackage{graphicx}

\usepackage{xcolor}
\usepackage{float}
\usepackage{titlesec}
\usepackage{setspace}
\usepackage{tcolorbox}
\usepackage[framemethod=TikZ]{mdframed}
\tcbuselibrary{skins}
\definecolor{grayblue}{rgb}{0.4, 0.4, 0.6}

\DeclareCaptionFormat{algoname}{\begin{tcolorbox}[colback=blue!1.2, boxsep=0pt, left=0pt, right=0pt, top=0pt, bottom=0pt, boxrule=0pt, arc=0pt, auto outer arc]\hrule height 1pt\vspace{1mm}\textbf{#1#2}#3\vspace{1mm}\hrule height 1pt\end{tcolorbox}}
\newmdenv[
  hidealllines=true,
  backgroundcolor=grayblue!3.2,
  skipabove=\baselineskip,
  skipbelow=\baselineskip,
]{coloredalgorithm}

\newcommand{\lizy}[1]{\textcolor{black}{#1}}
\begin{document}

\title{Optimization Techniques for Unsupervised Complex Table Reasoning via Self-Training Framework}

\author{
Zhenyu Li, Xiuxing Li, Sunqi Fan~\IEEEmembership{Member,~IEEE,}
Jianyong Wang,~\IEEEmembership{Fellow,~IEEE}
\thanks{Z. Li, S. Fan, and J. Wang are with the Department of
Computer Science, Tsinghua University, Beijing 100084, China.
E-mail: \{zy-li21, fansq20\}@mails.tsinghua.edu.cn,
jianyong@tsinghua.edu.cn.}
\thanks{X. Li is with University of Chinese Academy of Sciences, Beijing, China. E-mail: lixiuxing@ict.ac.cn.}
}

\markboth{Journal of \LaTeX\ Class Files,~Vol.~14, No.~8, August~2021}%
{Shell \MakeLowercase{\textit{et al.}}: A Sample Article Using IEEEtran.cls for IEEE Journals}

\IEEEpubid{0000--0000/00\$00.00~\copyright~2021 IEEE}


\maketitle

\begin{abstract}
Structured tabular data is a fundamental data type in numerous fields, and the capacity to reason over tables is crucial for answering questions and validating hypotheses. However, constructing labeled data for complex reasoning tasks is labor-intensive, and the quantity of annotated data remains insufficient to support the intricate demands of real-world applications. To address the insufficient annotation challenge, we present a self-training framework for unsupervised complex tabular reasoning (UCTR-ST) by generating diverse synthetic data with complex logic. Specifically, UCTR-ST incorporates several essential techniques: we aggregate diverse programs and execute them on tables based on a ``Program-Management" component, and we bridge the gap between programs and text with a powerful ``Program-Transformation" module that generates natural language sentences with complex logic. Furthermore, we optimize the procedure using  ``Table-Text Manipulator"  to handle joint table-text reasoning scenarios. The entire framework utilizes self-training techniques to leverage the unlabeled training data, which results in significant performance improvements when tested on real-world data. Experimental results demonstrate that UCTR-ST achieves above 90\% of the supervised model performance on different tasks and domains, reducing the dependence on manual annotation. Additionally, our approach can serve as a data augmentation technique, significantly boosting the performance of supervised models in low-resourced domains\footnote{The code and models are available on \url{https://github.com/leezythu/UCTR-ST}.}. 
\end{abstract}


\section{Introduction}
\IEEEPARstart{T}{abular} data is a widespread format for presenting information in the real world. This structure allows for the concise and efficient display of data. For instance, Wikipedia infoboxes utilize fixed-format tables to summarize relevant information with shared characteristics succinctly. Furthermore, tables are ubiquitous in various specific domains, such as scientific documents~\cite{wang2021semeval}, financial reports~\cite{zhu2021tat}, education~\cite{jauhar2016tabmcq}, and industry~\cite{katsis2021ait}. Recent years have witnessed the remarkable development of tabular reasoning, which has achieved tremendous success in various downstream application areas. The fact verification task~\cite{chen2019tabfact},~\cite{gupta2020infotabs} and the question-answering task~\cite{zhong2017seq2sql},~\cite{pasupat2015compositional} are two prevalent reasoning tasks that assess a model's ability to comprehend and interpret tabular data effectively. Table fact verification is a natural language inference task~\cite{bowman2015large} with evidence in structured forms. Given a table as evidence, the model is required to determine whether a textual hypothesis is ``supported'', ``refuted'', or ``unknown''. For table question answering, the model takes a table with a table-related natural language question as its input and returns the corresponding answer. Moreover, table-text reasoning tasks are introduced to better align with real-world requirements. The model must evaluate the table and related text concurrently to provide accurate judgments or answers. In summary, investigating tabular reasoning is crucial, as it facilitates more efficient techniques of the vast array of structured table resources that emerged on the web and in databases.

\begin{figure}
  \centering
  \includegraphics[width=1\linewidth]{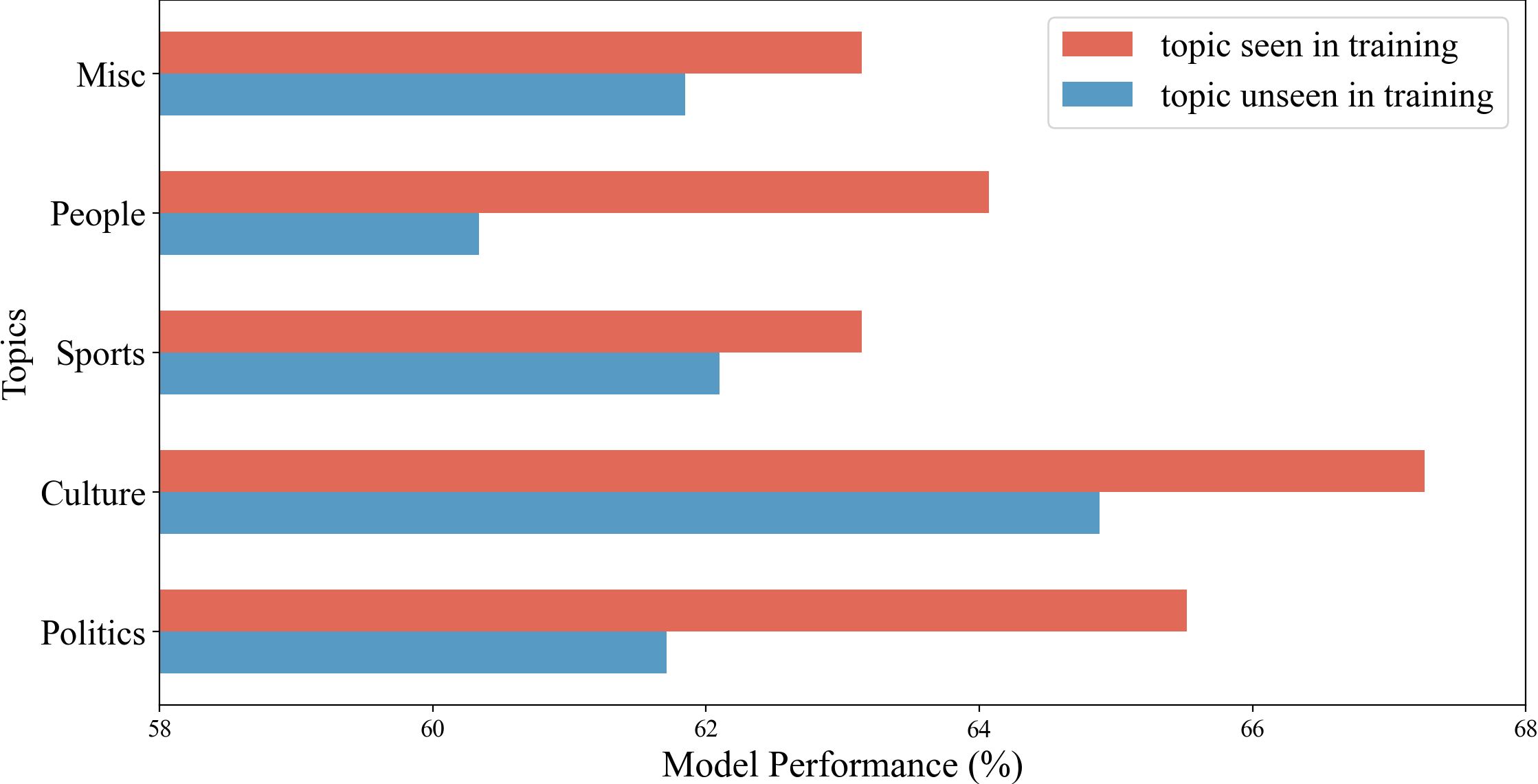}
  \caption{The previous study \cite{chemmengath2021topic} shows performance of models degrades dramatically on topics not seen during the training stage. }
  \label{performance drop}
\end{figure}

\IEEEpubidadjcol
The essential challenges of tabular reasoning involve accurately comprehending table structures and effectively capturing the relationships between table cells or between table cells and related sentences. Despite the tremendous success of pre-trained language models (PLMs) in textual reasoning tasks (e.g., textual entailment \cite{dagan2005pascal}, and question answering \cite{rajpurkar2016squad}), they primarily rely on free-form textual data for pretraining. Consequently, the considerable format discrepancy between free-form texts and structured tables seriously restricts effective table reasoning. To further alleviate the challenges that existed in tabular reasoning, an ever-growing tendency of adapting pre-trained models to tables has emerged \cite{wang2021tuta}, \cite{herzig2020tapas}, \cite{nassar2022tableformer}, \cite{yin2020tabert}. They explore diverse table-oriented model architectures \cite{arik2021tabnet}, \cite{deng2022turl}, \cite{eisenschlos2021mate} and pre-training objectives \cite{yu2020grappa}, \cite{liu2021tapex}, \cite{xie2022unifiedskg} to leverage the particular properties of the table better. Moreover, they investigate distinct serialization methods \cite{iida2021tabbie}, \cite{gong2020tablegpt} to linearize tables to sequences, attempting to eliminate the gap. These methods have achieved significant improvements over previous approaches \cite{choi2021ryansql}, \cite{wang2019learning}, \cite{hwang2019comprehensive}.

\IEEEpubidadjcol
However, the aforementioned methods are based on the assumption that adequate training data is available, which may not always be true. When human-annotated data is insufficient, these approaches might suffer considerable limitations and experience substantial performance deterioration. Additionally, Chemmengath et al. \cite{chemmengath2021topic} observe a significant decrease in the model's performance when encountering samples from topics not covered during the training stage. As demonstrated in Figure \ref{performance drop}, the model's performance declines markedly when exposed to previously unseen topics. 
To emancipate the limitations of the above assumptions and make the setting of tabular reasoning more in conformity with the actual scenarios, the unsupervised complex tabular reasoning setting has been proposed, which means reasoning on tables or a hybrid of tables and related text using complex logic with no manually annotated data available. These methods can be generally divided into two optimization directions: (1) Methods based on pre-training process reconstruction. These methods are designed as data-augmentation techniques with limited unsupervised performance. In addition, they always require a large pre-training corpus. Yu et al. \cite{yu2020grappa} pre-train the model on a large amount of question-SQL pairs, and Liu et al. \cite{liu2021tapex} show that synthetic SQL queries can provide a better model initialization. (2) Methods based on synthesizing human-like data through heuristics or data-to-text models. Eisenschlos et al. \cite{eisenschlos2020understanding} generate claims using context-free grammar (CFG) templates and counterfactual heuristics. Recently, Pan et al. \cite{pan2020unsupervised}  propose an unsupervised learning framework named MQA-QG. MQA-QG generates multi-hop questions for both the tabular and textual data, which is the most relevant work in this tendency. 

  \begin{figure}
  \centering
  \includegraphics[width=0.95\linewidth]{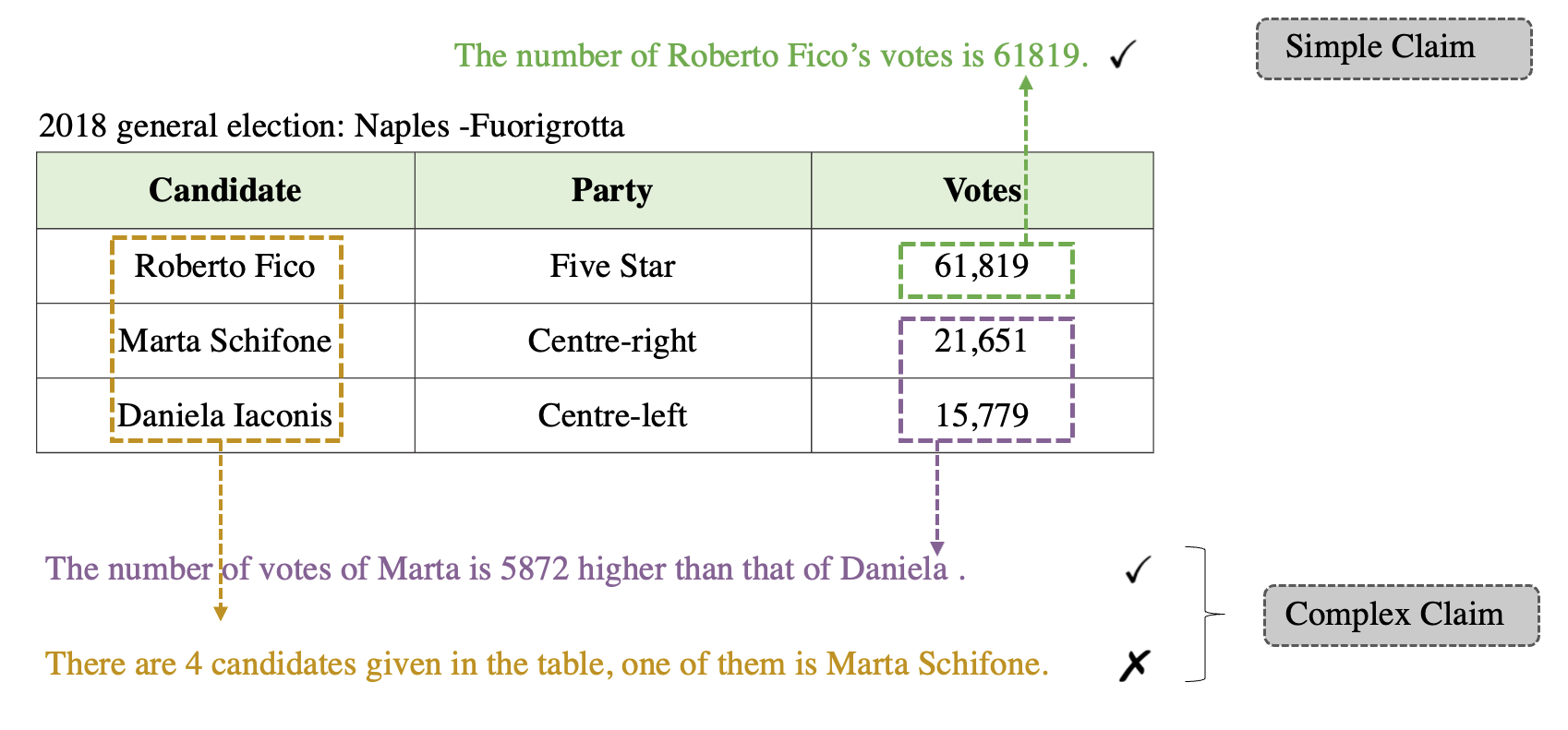}
  \caption{The comparison of simple claims and complex claims. A simple claim only involves a specific table cell, but a complex claim requires the annotator to consider the relationship among multiple cells.}
  \label{complex reasoning example}
\end{figure}

Nevertheless, several critical issues remain unsolved in the realm of unsupervised complex tabular reasoning: (1) Existing methods for data generation mainly use heuristics or shallow data-to-text methods (e.g., converting a row to a sentence). Thus, they can merely generate relatively simple instances shown in Figure \ref{complex reasoning example}, limiting the model's effectiveness on complex reasoning samples, which require a deep understanding of the semantics and logic relationships between multiple table cells. (2) Previous works only focus on a single scenario but cannot be expanded to other tabular reasoning tasks. This is because they design heuristics based on specific data characteristics or the form of the task, and these methods cannot be transferred to other tasks flexibly. Therefore, these models struggle to handle the complex and diverse scenarios encountered in the real world.

To address these issues, we introduce UCTR-ST (\textbf{U}nsupervised \textbf{C}omplex \textbf{T}abular \textbf{R}easoning using \textbf{S}elf-\textbf{T}raining), an advanced self-training framework designed explicitly for unsupervised complex tabular reasoning. More specifically, UCTR-ST primarily leverages a random sampling strategy to collect different types of programs. These programs consist of sequences of symbols that can be executed on tables, including SQL queries, logical forms, and arithmetic expressions, encompassing a wide range of reasoning types. Subsequently, we design a \lizy{``Program-Management"} module that generates program-answer pairs by leveraging numerous tables within the domain. To bridge the gap between the programs and natural language sentences, we develop a powerful \lizy{``Program-Transformation"} module based on generative language models that turns the programs into human-like natural questions or claims with complex logic. Since a table often occurs with its surrounding texts, UCTR-ST also defines a \lizy{``Table-Text Manipulator". It contains} two basic operators: Table-To-Text and Text-To-Table, to fuse information from table and text sources. Based on the combination of these components, UCTR-ST can handle question-answering and fact-verification tasks under both the homogeneous (table only) setting and the heterogeneous (hybrid table and text) setting, aiming for a unified framework. Figure \ref{framework} illustrates the progress that UCTR-ST generates a joint table-text reasoning sample through a SQL query, utilizing our foundational modules and operators. Experiment results show that UCTR-ST can generate diverse and human-like training samples with complex logic, which results in surprising unsupervised performance. Additionally, existing models pre-trained on our synthetic dataset significantly outperform the supervised model under the few-shot setting. Finally, we reconstruct the model's training process via the self-training framework, refraining from an exclusive reliance on synthetic data, which may not accurately represent the distribution of real-world data. Therefore, UCTR-ST can further improve the model's performance by effectively leveraging unlabeled realistic data. UCTR-ST employs the model initially trained on synthetic data to assign labels to the unlabeled real-world data, incorporating these pseudo-labeled samples into the training set. By repeating this process, the model achieves self-boosting, resulting in significant performance improvements. The experimental results indicate that UCTR-ST effectively utilizes unlabeled data to achieve remarkable enhancements compared to UCTR\lizy{, a basic version of UCTR-ST that doesn't use the self-training technique}. Notably, we also discover that UCTR-ST can considerably enhance fully-supervised performance in low-resource domains. 
Our main contributions can be summarized as follows:

\begin{itemize}
  \item To the best of our knowledge, this is the first study exploring  a unified unsupervised complex tabular reasoning framework.
 
\item We propose a novel and effective framework by leveraging program generation and conversion modules to cope with unsupervised complex reasoning.
\item We further design novel ``Table-to-Text" and ``Text-to-Table" operators in UCTR to handle joint table-text reasoning scenarios.
  \lizy{\item In order to better utilize the information from the distribution of unlabeled real data, we employ self-training techniques to enable the model to self-boost, which demonstrates general effectiveness of UCTR-ST across various tasks.}
  \item Comprehensive experiments show that UCTR \lizy{and UCTR-ST} can significantly benefit tabular reasoning systems under unsupervised, few-shot, and even supervised settings.

\end{itemize}

\section{Preliminary}
In this section, we start with introducing the background knowledge of the tabular reasoning task. In particular, we first present related basic concepts and then formalize the tabular reasoning task. Afterwards, since this paper aims to tackle the unsupervised scenario, we give a brief overview of the primary unsupervised data generation approaches.

\subsection{Tabular Reasoning.}
We first define some basic concepts related to the tabular reasoning task.

\textbf{Table.} The structure of a table can be very flexible, and we can divide tables into various categories according to their different formats \cite{dong2022table}. Among them, relational tables are the most commonly used. For a relational table $T$ with n rows $\{r_1,...,r_n\}$, each row can be seen as a record, with columns as the corresponding attributes. 

\textbf{Context.} In most cases, there are related paragraphs $P$ surrounding a table as its context. These texts always describe the table's contents or contain supplementary information. Some tabular reasoning tasks require the model to consider not only the evidence from tables, but also the evidence in textual form. Reasoning on heterogeneous data is more realistic and challenging.

\textbf{Tabular Reasoning}. In this paper, we define tabular reasoning as reasoning tasks on tabular evidence or joint table-text evidence. Specifically, we use two tasks: tabular fact verification and tabular question answering, to evaluate a model's reasoning ability. We can formalize the task as a mapping from the evidence and a natural language sentence $L$ to an output $O$. The basic mapping can be written as:
\begin{equation}
    f(T,L) \rightarrow O \label{eq:tabular reasoning}
\end{equation}
 If the evidence consists of both a table and its related text, the mapping can be extended as:
 \begin{equation}
    f(T,P,L) \rightarrow O\label{eq:tabular reasoning2}
\end{equation}
 
We present detailed explanations of the equation for each specific task below.

\textbf{Tabular Fact Verification}. Given a table as evidence, tabular fact verification requires the model to judge whether the evidence supports, refutes a natural language claim, or it's unknown. That is, $O\in\{Supported, Refuted, Unknown\}$ in Equation \ref{eq:tabular reasoning} and \ref{eq:tabular reasoning2}. It is similar to traditional fact verification task on textual data \cite{thorne2018fever}, except the evidence format.

\textbf{Tabular Question Answering}. Similarly, tabular question answering is a migration of the traditional question answering task from textual data to tabular data.
Its output is always a specific answer inferred from the evidence.

\textbf{Complex tabular reasoning}. We define complex tabular reasoning as the reasoning process of considering multiple table cells and understanding their logical relationships to infer the correct answer. In contrast, simple tabular reasoning only involves a single table cell, as depicted in Figure \ref{complex reasoning example}. Simple reasoning tasks are easier to solve, since models are good at learning associations between surface texts.

\textbf{Program}. A program is an executable sequence of symbols \cite{pi2022reasoning}, such as a SQL query. Unlike natural language texts, programs have strict grammar rules with no ambiguity and have definite execution results.
As a related concept, ``program context'' refers to an environment where a program is applied. The variables used in the program are also sampled from the context. For example, tables are the corresponding context for SQL queries. 
Besides, we refer to a ``program executor'' as an automated tool that executes a program within the context, such as a SQL executor. We can use the program executor as a black box, whose input is a program and program context, and output is the execution result.

Section \ref{program design} elaborates on the types of programs we used in this paper.

\subsection{Unsupervised Data Generation.}
Supervised models tend to show powerful results in an ideal environment where sufficient high-quality data is available. Unfortunately, we often face situations with a limited amount of labeled data or no labeled data in the real world, under which the model's performance suffers a severe decline inevitably. This dilemma leads to the research direction of unsupervised data generation, aiming to synthesize human-like training instances \cite{judea2014unsupervised}.

Formally, for tabular reasoning tasks, supervised models assume labeled training data $X=\{(t_1,p_1,l_1,o_1),\cdots,(t_i,p_i,l_i,o_i),\cdots,(t_n,p_n,l_n,o_n)\}$, where $n$ is the number of training instances. But under unsupervised settings, we only have $X = \{(t_1,p_1),\cdots,(t_i,p_i),\cdots,(t_n,p_n)\} $ as available information, where $t_i$, $p_i$, $l_i$, and $o_i$ are an unlabeled table, the related text, a natural language question/claim, and the corresponding golden label, respectively. The data generation method tries to reconstruct a synthetic training dataset $X^{'} = \{(t_1,p_1,l^{'}_1,o^{'}_1),\cdots,(t_i,p_i,l^{'}_i,o^{'}_i),\cdots,(t_n,p_n,l^{'}_n,o^{'}_n)\} $ using these raw tables and texts. Based on this synthetic dataset, supervised models can be applied successfully.

\lizy{However, the distribution of the generated data in the above manner may have a significant gap from the distribution of questions/claims from real users. Therefore, we can adopt a relaxed but more practical unsupervised data generation setting: we have $X = \{(t_1,p_1,l_1),\cdots,(t_i,p_i,l_i),\cdots,(t_n,p_n,l_n)\} $ as available information. In the subsequent experiments, we demonstrate that, guided by the information of real questions/claims, the model can achieve better performance on real test data.}

\subsection{Program Design.}\label{program design}

In this paper, we adopt three types of programs: logical forms, SQL queries and arithmetic expressions. We depict examples of their forms and execution results on a table in Figure \ref{fig:types of programs}. Among them, logical forms are used for fact verification tasks, while SQL queries and arithmetic expressions are used for question answering tasks. Due to the variety of logic operators and flexible structure, the programs can cover most types of logic used in tabular scenarios. We give more detailed explanation of each program type below:

\textbf{SQL Queries}. SQL is standard language for managing data, which is widely used in relational databases. SQL supports many manipulations like query, insert, update, and delete, but we only need SQL queries for our reasoning setting. 
In most cases, you can query any content you want to know from the table through one or more SQL queries. Specifically, the SQL queries support the following reasoning types (conditions): $equivalence\ (=)$, $comparison\ (\textgreater, \textless, order\ by, max, min)$, $counting\ (count)$, $sum\ (+)$, $diff\ (-)$ and $conjunction\ (and)$.

\textbf{Logical Forms}. Though SQL queries are powerful, they cannot be directly used on tabular fact verification tasks. So in our framework, we generate factual claims based on logical forms specifically. A Logical form is a symbolic formulation that can be executed on database tables to judge the truthfulness of the inner logic. Logical Forms can also support most common reasoning types such as: $count$, $superlative$, $comparative$, $aggregation$, $majority$, $unique$, and $ordinal$. For example, in the logical form depicted in Figure \ref{fig:types of programs}, $argmax$ returns the row with the max value under the specified column, and $hop$ extracts the value under a specified column for an input row. Finally, $eq$ judges whether the two arguments are equal. Due to the space limitation, we refer readers to \cite{chen2020logic2text} for a complete list of the operations. Due to the limited space, we refer readers to \cite{chen2020logic2text} for the full list of operators.

\textbf{Arithmetic Expressions}. Arithmetic expressions can be used to express complex arithmetic operations. As shown in Figure \ref{fig:types of programs}, an arithmetic expression consists of a sequence of operations. Arithmetic expressions support 6 mathematical operations: $add$, $subtract$, $multiply$, $divide$, $greater$, $exp$ and 4 table aggregation operations $table\_max$, $table\_min$, $table\_sum$, $table\_average$. We refer readers to \cite{chen2021finqa} for more detailed illustrations.

\section{Framework}

\begin{figure*}
  \centering
  \includegraphics[width=0.879\linewidth]{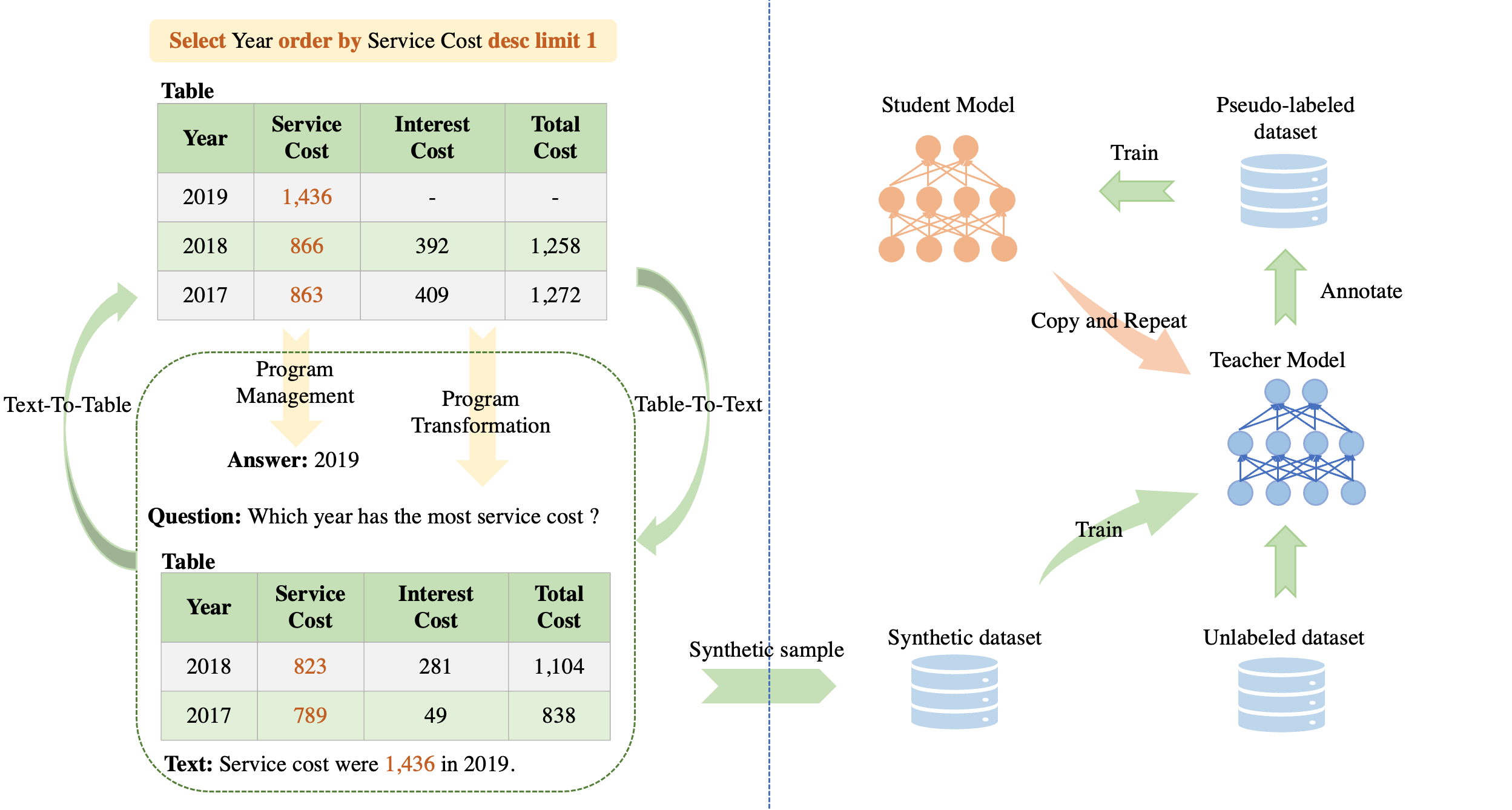}
  \caption{Illustration of our framework. The left part depicts how we generate synthetic samples (enclosed in the dashed box). Specifically, the Table-To-Text operator focuses on splitting the original table into a sub-table and a generated sentence and then building a joint table-text reasoning sample based on the basic modules. The Text-To-Table operator adopts a similar procedure  but aggregates information from the original table and text to form an expanded table. The right part show how we apply the self-training technique. In each iteration, the teacher model infers on the unlabeled data to generate pseudo-labeled data, which is then used to train a better student model.}
  \label{framework}
\end{figure*}

In this section, we present our proposed \lizy{self-training framework for unsupervised complex tabular reasoning, UCTR-ST. UCTR-ST uses three essential modules to generate human-like data for various tabular reasoning scenarios: Program-Management, Program-Transformation, and Table-Text Manipulator. The left part of Figure \ref{framework} shows how UCTR-ST generates joint table-text reasoning instances using SQL queries on a table from the TAT-QA dataset. Note that the Table-Text Manipulator consists of two operators: Table-To-Text and Table-To-Text, corresponding to two different data generation methods. The right part of the figure depicts the self-training process. We show more details of each part of the workflow below. }

\subsection{Table Splitting.} 
As shown in the left part of Figure \ref{framework}, given a raw table, the table splitting generation method first executes a program based on the Program-Management module and gets an answer. Note that not all table cells affect the final output, and we define the cells involving the reasoning process as ``highlighted cells." Then the Table-To-Text operator selects one highlighted cell, and transforms the row where the cell is located into a natural language text, keeping the rest of the rows as a sub-table. Additionally, the Program-Transformation turns the program into a question with the same meaning.
In this way, we successfully synthesize a training instance $(t,p,l)\rightarrow o$ requiring evidence from both a table and its related text.

\subsection{Table Expansion.~} 
Table expansion can be regarded as an inverse process of table splitting. The table splitting method synthesizes joint table-text reasoning instances from only tables, while the table expansion method tries to integrate information from the original texts surrounding the table.
Specifically, the table expansion method first finds the relevant sentences and then uses the Text-To-Table operator to transform essential information of the sentences into tabular form. If the generated table shares the same row name or the same column name with the original table, they can be integrated into a new expanded table. Afterwards, UCTR-ST can apply the Program-Management and Program-Transformation techniques on this expanded table as in the table splitting method. Finally, we synthesize a joint table-text reasoning instance with evidence from the original table and text. 

Table-To-Text and Text-To-Table operators are designed for joint reasoning on heterogeneous data. For table-only scenarios, we can follow the same procedure but just use the Program-Management and Program-Transformation modules. Thus, UCTR-ST can become a unified framework that can cope with both homogeneous and heterogeneous scenarios.

\subsection{Tabular Reasoning Models.} 
Although researchers have designed different model structures for various tasks, the mainstream methods share the same paradigm as follows:
\begin{equation}
\begin{aligned}
       e_i &= Encoder(t_i,p_i,l_i)\\
       \theta_{model} &= \mathop{\arg\min}\limits_{\phi}L(Classifier(e_i),o_i)
\end{aligned}
\end{equation}
where $t_i$, $p_i$ , and $l_i$ represent the table, paragraph and natural language sentence in a sample. $o_i$ is the golden label of the sample. $L$ is the loss function, and $\theta$ is the model parameters. We first get a joint representation based on an encoder like BERT \cite{devlin2018bert}, then a designed classifier is applied to the representation to get predicting results. We optimize the model's parameters using gradient descent techniques during training. In experiments, we use the representative model on each task as the supervised baseline.

\subsection{\lizy{Self-Training}}
\lizy{Self-training is a popular technique in semi-supervised machine learning, which involves training a model with a small set of labeled data and a larger set of unlabeled data. But there are relatively few works that use the self-training technique in completely unsupervised scenarios. In this paper, we recognize synthetic samples as the existing labeled data and the samples in the training set as the unlabeled data that conforms to the natural distribution. }

\lizy{Specifically, we train an initial model based on the synthetic data and then make predictions on the unlabeled dataset. The examples for which the model makes predictions are then added to the labeled dataset and used to re-train the model. This iterative process can improve the robustness of the model and also the quality of the labeled dataset with each iteration. Eventually, the model can converge to a better performance.
}

\section{Methodology}
In this section, we present the workflow of the UCTR-ST framework, along with more detailed information on each technique used, demonstrating how they can achieve our ultimate goals: i) generating human-like training instances with complex logic, ii) being able to handle various table reasoning scenarios, and iii) achieving better testing performance with unlabeled real-world data. Specifically, we first formalize the modules and their corresponding functions included in each technique, and then provide specific explanations for how we apply programs and training models.

\subsection{Program-Management.}\label{Synthetic-Supervision.}
This component aims to aggregate diverse programs with complex logic and execute them on tables. Formally, we can define these two procedures as follows: 

\textbf{Program Generation.}\ 
Given raw tables in a specific domain, this procedure retrieves program templates from a template pool and applies these templates to tables to get executable programs:
\begin{equation}
    f(T) \rightarrow Prog
\end{equation}

\textbf{Program Execution.}\ 
The function of our Program Execution component is the same as stated in the preliminary. Given a table and a program as input, the executor returns the execution result:
\begin{equation}
    f(T,Prog) \rightarrow O
\end{equation}
Programs in each type rely on a specific executor.
We give a more detailed explanation for these programs in section \ref{program collection} and section \ref{implementation details}.

\subsection{Program-Transformation}

The advantage of programs compared to natural language is that there is no ambiguity and they can give definite execution results according to the grammar rules so that we can get concise program-answer pairs. However, different types of programs follow different grammar rules, and there is a huge gap in the surface form between a program and a natural language sentence. Therefore, we design a program-transformation component for mapping different programs of different types into a unified natural language format. Formally, it can be regarded as a mapping function as follow:
\begin{equation}
    f(P) \rightarrow L
\end{equation}
where $P$ is a program and $L$ is the corresponding natural language sentence with the same meaning. 

\subsection{Table-Text Manipulator}


\textbf{Table-To-Text.}\
This operator converts a table into a sub-table and a generated sentence. Formally, the function is defined as:
\begin{equation}
    f(T) \rightarrow T_{sub},S
\end{equation}
Specifically, we follow the implementation of  ``DescribeEnt'' operator in \cite{pan2020unsupervised} to transform a row into a natural language sentence, and more advanced models \cite{kale2020text}, \cite{su2021plan} can also be used here. Additionally, we add a filtering step. That is, if important information in the table is missing from the generated sentence, we will discard it.

\textbf{Text-To-Table.}\
As an inverse process of Table-To-Text, the function of Text-To-Table can be written as:
\begin{equation}
    f(T,P) \rightarrow T_{expand}
\end{equation}
Actually, text-to-table is a recently proposed task for information extraction \cite{wu2021text}. But current techniques do not support integrating text information into existing tables. So a filtering step is also needed here. We first use row names to filter possible useful sentences, and then apply a text-to-table model proposed in \cite{wu2021text} to get a generated table with only one record. Finally, we integrate this record into the original table to form an expanded table.

\subsection{How we collect program templates}\label{program collection}

\begin{figure*}
  \centering
  \includegraphics[width=0.8\linewidth]{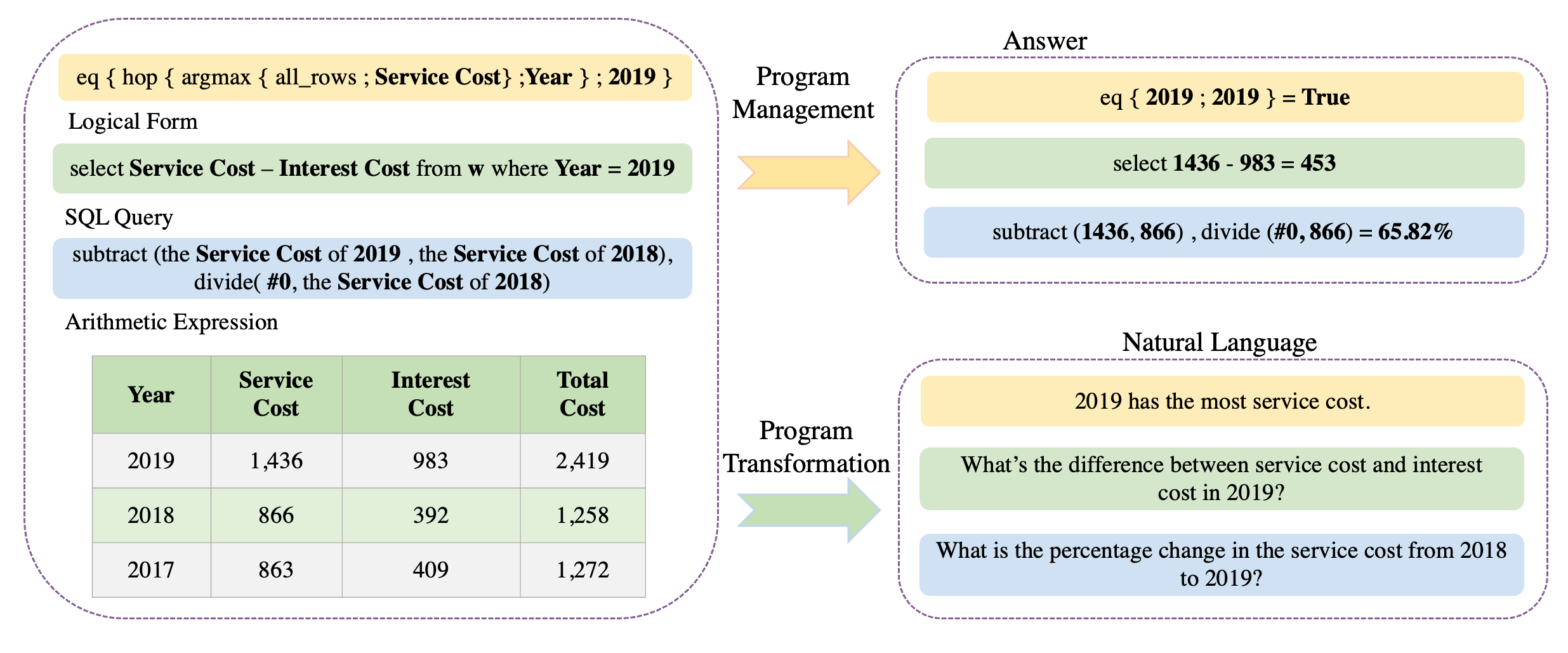}
  \caption{Examples of three types of programs we used in this work: logical forms, SQL queries, and arithmetic expressions. The Program-Transformation module transforms a logical form into a claim and transforms the other two types of programs into questions.}
  \label{fig:types of programs}
\end{figure*}

The three types of programs (SQL queries, logical forms, and arithmetic expressions) are essential parts of UCTR-ST. In this section, we explain the necessity of each type of program and show how we collect templates and apply them on tables to get program-answer pairs. We depict examples of each type of program in Figure \ref{fig:types of programs}.

For SQL query templates collection, we follow the implementation in \cite{liu2021tapex}, using templates extracted from SQUALL\cite{shi2020potential}. SQUALL is a dataset consisting of question-SQL pairs with manual alignments. One example template from SQUALL is as follows:
\begin{center}
    $select \ \textbf{c1}\ from\ \textbf{w}\ order\ by\ \textbf{c2\_number}\ desc \ limit\ 1$
\end{center}
where $w$ represents the table, and $c1$ and $c2$ correspond to the first and the second column. $\_number$ indicates that the data of this column is numerical.
These placeholders allow the template to migrate to other tables conveniently.

For logical forms, we use the LOGIC2TEXT dataset proposed by Chen et al. \cite{chen2020logic2text}. LOGIC2TEXT consists of a large number of claim-program pairs, covering most common logic types such as count, comparative, aggregation etc. We directly sample program templates from it. Here is an example of the template:
\begin{center}
    $eq \ \{\ hop\ \{\ filter\_eq\ \{\ all\_rows\ ;\ \textbf{c1}\ ;\ \textbf{val1}\ \}\ ;\ \textbf{c2} \} ; \textbf{val2} \}$
\end{center}
where $val1$ and $val2$ are cell values from the first and second columns. $eq$, $hop$, and $filter\_eq$ are defined operations. Specifically, $filter\_eq$ returns rows that satisfy the constraints.

Arithmetic operations are very common in some specific tabular reasoning scenarios (like financial and scientific). Although SQL can implement most types of operations, expressing arithmetic operations using SQL always results in very long sequences. Thus, we adopt arithmetic expressions for tabular reasoning tasks involving arithmetic operations as our programs. Specifically, we collect templates of arithmetic expressions from the Finqa dataset proposed in \cite{chen2021finqa}. The original form of a template is as follows:

\begin{center}
    $subtract(\textbf{val1}, \textbf{val2}), divide(\textbf{\#0}, \textbf{val2})$
\end{center}

where $\#0$ denotes the result from the first $subtract$ step. But the original form doesn't contain the information of the row's name or column's name, so we further replace $vali$ with $col\_name\ of\ row\_name$, where $col\_name$ and $row\_name$ are the column's name and row's name corresponding to $vali$. 
For more details about the arithmetic operations please refer to \cite{chen2021finqa}.

\subsection{How to apply program templates}
 We call the column names, and cell values involved in the program template as column-placeholders and value-placeholders, respectively. To apply these programs to a new table, we need to fill these placeholders with variables from the table. Here we adopt the random sampling strategy with type constraints for program sampling. Specifically, we first populate the column-placeholders by randomly sampling from the columns of the new table. Afterwards, for each column, we randomly sample the values in it to populate the value-placeholders. Besides, if the column-placeholder specifies a data type (e.g., number, string), we only sample from columns that match that type.

Take the logical form above as an example. The original program template is:
\begin{center}
    $eq \ \{\ hop\ \{\ filter\_eq\ \{\ all\_rows\ ;\ \textbf{c1}\ ;\ \textbf{val1}\ \}\ ;\ \textbf{c2} \} ; \textbf{val2} \}$
\end{center}
For a new table $T$, we first fill in the column-placeholders:
\begin{center}
    $\{c1,c2\}\leftarrow Random\_Sample(T.columns,data\_type)$
\end{center}
Then we fill in each value-placeholders:
\begin{center}
    $val1\leftarrow Random\_Sample(c1.values)$ \\
    $val2\leftarrow Random\_Sample(c2.values)$
\end{center}
In practice, for logical form templates with a format $func\ \{\ arg1\ ;\ arg2\ \}$, in which $func$ is the root operator, $arg1$ is a complex sub-template, and $arg2$ is a single value.
We first apply sampling on $arg1$ and execute it. Then we can determine the value of $arg2$ based on the execution result and the root operator to obtain a true/false claim.

In summary, this mapping strategy keeps the internal relationship of the variables in the original program. Moreover, this strategy is naturally suitable for evidence-based reasoning tasks since the values sampled during the process are exactly the evidence associated with the synthetic instance. Notably, if the execution result is empty, we discard this program.

\begin{algorithm}[ht] 
\begin{coloredalgorithm}
  \caption{Data Generation Procedure}
  \label{alg1}
  \begin{algorithmic}[1]
    \setstretch{1}
    \Require Table-text dataset $D_t$, program template dataset $D_m$
    \Ensure Target dataset $D$ consisting of ($L$, $T$, $P$, $O$) pairs 
    \State $D \leftarrow []$
     
    \For{table, paragraph in $D_t$}
        \State $table^{exp} \leftarrow \text{TextToTable}(table, paragraph)$
        
        \For{template in $D_m$}
            \State $prog \leftarrow \text{Sampling}(table, template)$
            \State $prog^{exp} \leftarrow \text{Sampling}(table^{exp}, template)$
            
            \State $ans \leftarrow \text{Executor}(table, prog)$
            \State $ans^{exp} \leftarrow \text{Executor}(table^{exp}, prog^{exp})$
            
            \If {$ans\ \text{or}\ ans^{exp}\ \text{is}\ \text{empty}$}
                \State \textbf{continue}
            \EndIf
            
            \State $table^{sub},\ sentence \leftarrow \text{TableToText}(table)$
            
            \State $NL \leftarrow \text{Transformation}(prog)$
            \State $NL^{exp} \leftarrow \text{Transformation}(prog^{exp})$
            
            \State $D_t.\text{append}((NL, table^{sub}, sentence, ans))$
            \State $D_t.\text{append}((NL^{exp}, table, paragraph, ans^{exp}))$
        \EndFor
    \EndFor  
    
  \end{algorithmic}
\end{coloredalgorithm}
\vspace{-1em} 
\end{algorithm}

\begin{algorithm}[ht] 
\begin{coloredalgorithm} 
\setstretch{1}
  \caption{Self Training Algorithm}
  \label{alg2}
  \normalsize
  \begin{algorithmic}[1]
     \Require Synthetic dataset $D_{s} = \{(l^{s}_i,t^{s}_i,p^{s}_i,o^{s}_i)\}$ generated from Algorithm 1 as labeled data; Unlabeled realistic dataset $D_{r} = \{(l^{r}_i,t^{r}_i)\}$.
     \Ensure Target model parameter $\theta_{\text{final}}$
     
    \Statex \textbf{Step 1.} Fine-tune teacher model $\theta_{\text{tea}}$ on labeled synthetic data
    \State $\theta_{\text{tea}} = \arg\min_{\theta}\ \text{loss} \{(l^{s}_i,t^{s}_i,p^{s}_i,o^{s}_i)\}$ 
     
    \While {not converged}
        \Statex \textbf{Step 2.} Generate pseudo-label for unlabeled real dataset
        \State $(p^{r}_i,o^{r}_i) \leftarrow f((l^{r}_i,t^{r}_i);\theta_{\text{tea}})$
         
        \Statex \textbf{Step 3.} Merge the labeled synthetic data and pseudo-labeled realistic data into a union of data
        \State $D_{u} = \{ (l^{s}_i,t^{s}_i,p^{s}_i,o^{s}_i) \} \cup \{ (l^{r}_i,t^{r}_i,p^{r}_i,o^{r}_i) \}$
         
        \Statex \textbf{Step 4.} Fine-tune student model $\theta_{\text{stu}}$ on the union of labeled synthetic data and pseudo-labeled real data
        \State $\theta_{\text{stu}} = \arg\min_{\theta}\ \text{loss} \{(l^{u}_i,t^{u}_i,p^{u}_i,o^{u}_i)\}$
         
        \Statex \textbf{Step 5.} Update the teacher model
        \State $\theta_{\text{tea}} \leftarrow \theta_{\text{stu}}$
    \EndWhile
  \end{algorithmic}
\end{coloredalgorithm}
\vspace{-1em} 
\end{algorithm}

\subsection{How to train the generative model.}
For the program-transformation, there are few works on converting a program to a natural language sentence. This paper tackles this problem based on generative language models.
For logical forms, we directly use the fine-tuned GPT-2 \cite{radford2019language} model on the Logic2Text \cite{chen2020logic2text} dataset. For SQL queries and arithmetic expressions, we fine-tune a BART \cite{lewis2019bart} model ourselves on SQUALL \cite{shi2020potential} and Finqa \cite{chen2021finqa}, respectively. These three datasets contain program-NL training pairs for each type of program.
Here we also briefly introduce generative models (e.g., GPT-2 and BART). They are transformer-based models pre-trained on a large corpus of text in an unsupervised manner and have been demonstrated to be very effective on machine translation tasks. For more details please refer to \cite{radford2019language}, \cite{lewis2019bart}. In our work, we recognize converting a program to a natural language sentence as a translation task, that is, translating a program into a sentence. Specifically, we fine-tune generative models in an end-to-end manner:
\begin{equation}
   L = Generative\_Models ( Prog )
\end{equation}
Here the generative models can be GPT-2, T5 and BART, etc.

In summary, we first collect a set of diverse program templates and then apply them to new tables by random sampling variables from the new context to get valid programs. Finally, we convert these programs into human-like reasoning instances using four basic components.

We depict the overall data synthesizing procedure using both table splitting method and table expansion method in Algorithm \ref{alg1}, and the self-training procedure in Algorithm \ref{alg2}.

\vspace{-.5em}
\section{PERFORMANCE EVALUATION}\label{experiments}\label{sec:experiment}

\subsection{Dataset and Evaluation.}
\textbf{Datasets.} To test the effectiveness of our UCTR-ST framework, we apply it in various settings. We conduct extensive experiments on four representative benchmarks:  FEVEROUS \cite{aly2021feverous}, TAT-QA \cite{zhu2021tat}, WiKiSQL \cite{zhong2017seq2sql}, and SEM-TAB-FACTS \cite{wang2021semeval}. The four datasets cover fact verification and question answering tasks under table-only and table-text reasoning scenarios, in general and specific domains. Here we give a brief introduction to each dataset. FEVEROUS is a dataset for fact verification over evidence from sentences and tables within Wikipedia. TAT-QA is also built on hybrid data but aims for question answering task. Additionally, its evidence is extracted from real-world financial reports. WiKiSQL consists of examples of questions and SQL queries over tables from Wikipedia. SEM-TAB-FACTS is a fact verification dataset with evidence in tabular form, and the tables are from scientific articles.
Statistics of these datasets are shown in Table  \ref{datasets}.

\begin{table*}[ht]
  \centering
  \caption{dataset statistics of FEVEROUS, TAT-QA, WiKiSQL and SEM-TAB-FACTS.}
  \begin{tabular}{ccccc} \toprule
  Dataset &  \ Domain & \ Total Samples  & \ Evidence Type & Label/Question Types\\ \midrule
\multirow{2}{*}{FEVEROUS}    & \multirow{2}{*}{Wikipedia}  & \multirow{2}{*}{87,026}  & 34,963 sentences,\ 28,760 tables  &49,115 Supported,\ 33,669 Refuted\\ 
&&&24,667 combined&4,242 NEI\\  \midrule

\multirow{2}{*}{TAT-QA}    & \multirow{2}{*}{Finance} & \multirow{2}{*}{16,552} & 7,431 tables,\ 3,902 sentences&9,211 Span/Spans,\ 377 Counting  \\ 
&&&5,219 combined&6,964 Arithmetic\\ \midrule

\multirow{2}{*}{WIKISQL}    & \multirow{2}{*}{Wikipedia} & \multirow{2}{*}{80,654} & \multirow{2}{*}{24,241 tables}&43,447 What,\ 5,991 How many \\ 
&&&&5,829 Who,\ ...\\ \midrule

\multirow{2}{*}{SEM-TAB-FACTS}   & \multirow{2}{*}{Science} & \multirow{2}{*}{5,715} &\multirow{2}{*}{ 1,085 tables} &3,342 Supported,\ 2,149 Refuted\\ 
&&&&224 Unknown \\
 \bottomrule
  \end{tabular}
  \label{datasets}
\end{table*}

\textbf{Evaluation protocol.} 
There are different evaluation metrics for different benchmarks. Typically, the pipeline of a model on FEVEROUS consist of retrieving stage and reasoning stage. In the first stage, the model retrieves sentences and table cells related to a claim from Wikipedia. Then in the second stage, the model judges whether the claim is supported, refuted, or there is not enough information (NEI) based on the evidence. So for FEVEROUS, the metrics are label accuracy and FEVEROUS score. Label accuracy measures the proportion of the number of correct labels predicted by the model to the total number. FEVEROUS score is a more strict metric that considers both the retrieving stage and the reasoning stage. For a sample, only when both the retrieved evidence set and the predicted label are correct is the prediction considered correct. Since the retrieving stage is not the focus in our paper, we directly use the retriever proposed in \cite{aly2021feverous} as our first-stage model and only experiment with the reasoning stage. Notably, we train the reasoning model on the golden evidence set rather than the retrieved evidence set, as the latter contains much noise. The metrics to measure performance on TAT-QA are Exact Match (EM) and numeracy-focused F1 score \cite{li2016dataset}. For WiKiSQL, the evaluation metric is denotation accuracy, which measures how many predicted answers are equal to the ground-truth answers. For SEM-TAB-FACTS, we adopt the standard 3-way micro F1 from the original paper. This metric evaluates whether claims are classified as Supported, Refuted, or Unknown.
\subsection{Implementation Details.} \label{implementation details}
For the ``Program-Transformation'' module, we choose the appropriate program type according to the setting of a task and the reasoning ability it requires. Specifically, we apply logical forms on FEVEROUS and SEM-TAB-FACTS tasks during the data generation procedure to generate claims with complex logic, and apply SQL queries on WiKiSQL. For TAT-QA, we apply both SQL queries and arithmetic expressions. As shown in Table \ref{datasets}, there are various reasoning types in TAT-QA. We use SQL queries to handle the Span/Spans type and use arithmetic expressions for the Counting and Arithmetic type. The tables we use to generate synthetic data are from the original datasets.
 Finally, we get 79,856, 23,933, 27,365, 4,071 synthetic samples for FEVEROUS, TAT-QA, WiKiSQL and SEM-TAB-FACTS, respectively. 
 
 As for model training, the entire process is based on a self-training framework. After each iteration, we either add the generated pseudo-labeled data to the original synthetic dataset or only use the generated pseudo-labeled data, depending on the performances on the dev set. We select representative models on the four benchmarks as our baselines. Specifically, we adopt the models in the original papers of FEVEROUS \cite{aly2021feverous} and TAT-QA \cite{zhu2021tat}. Since they achieve good results with reproducible codes. For WiKiSQL, we use the current state-of-the-art model TAPEX \cite{liu2021tapex}. For SEM-TAB-FACTS, we use a representative model, TAPAS \cite{herzig2020tapas}. Section \ref{results} shows more details of the models used on each benchmark. 
In experiments, we follow the implementation in \cite{malon2021team} on FEVEROUS, only predicting  the ``Supported'' or ``Refuted'' label, since the ``NEI'' label occupies a tiny proportion of the dataset.
Besides, we also evaluate models under a few-shot setting, where we assume only 50 human-labeled samples are available. The 50 samples are randomly selected from the original training set.

The executor for SQL queries is sqlite3 \footnote{https://docs.python.org/3/library/sqlite3.html}. For logical forms and arithmetic expressions, we utilize the executor proposed in \cite{chen2020logic2text} and \cite{chen2021finqa},  respectively. Experiments are conducted with 4 GeForce RTX 3090 graphics cards.

\subsection{Results Analysis. }\label{results}
Table \ref{unsuper on tat}, \ref{unsuper on feverous}, and \ref{unsuper on semeval} summarize the unsupervised and few-shot results on three benchmarks. In this section, we analyze the effectiveness of our self-training framework for unsupervised complex tabular reasoning (UCTR-ST) compared to supervised baselines. We first give illustrations of the supervised and unsupervised models we use. Representative supervised models are as follows:

(1) TAGOP is a strong supervised model designed for TAT-QA. It first tags relevant table cells and text spans and then reasons over these elements using a set of predefined operators. Text-Span only and Table-Cell only are two weak supervised baselines that adopt the same architecture as TAGOP, but they focus on textual evidence or tabular evidence only.

(2) Full baseline is the baseline model proposed in \cite{aly2021feverous} that consists of a retriever module that retrieves relevant table cells and sentences from Wikipedia and a verdict predictor that predicts a label. As mentioned above, since we only focus on the reasoning stage, we assume golden evidence is available when testing label accuracy. When testing the FEVEROUS score, we use the trained retriever in the original paper for a fair comparison. The Sentence-only baseline and Table-only baseline are two weak supervised models trained only on sentences or tables.

(3) TAPAS is a popular tabular reasoning model, using joint pre-training of textual and tabular data. It uses special positional embeddings to encode table structures and shows promising performance on fact verification and question answering tasks. We apply TAPAS on both TAT-QA and SEM-TAB-FACTS. The result on TAT-QA is from \cite{zhu2021tat}. For SEM-TAB-FACTS, we follow the method in \cite{gautam2021volta} to fine-tune TAPAS. 

(4) TAPEX is a generative pre-trained model that is pre-trained on a large SQL query-answer corpus to imitate a neural SQL executor, and it produces state-of-the-art results on the WiKiSQL dataset. We also use it as an unsupervised model to see how much the synthetic corpus can help the model cope with real questions. We evaluate the officially released tapex-base models and get the corresponding results on development and test sets.

 We compare our UCTR-ST framework with the following unsupervised models:

(1) Random is a naive baseline used for FEVEROUS and SEM-TAB-FACTS, selecting a label randomly. Since these two tasks are essentially multi-classification tasks, this baseline shows how much performance a model should at least achieve. Notably, the ``NEI'' label in FEVEROUS only occupies a tiny proportion, so we only predict the ``Supported'' or ``Refuted'' label in practice.

(2) MQA-QG is also an unsupervised data generation method, which is the most relevant work to ours. Though it is initially designed for multi-hop question generation, we make some modifications to fit it on these benchmarks. Specifically, MQA-QG finds a bridge entity that connects the table and related text, then turns the row containing the bridge entity into a describing sentence using a $DescribeEnt$ operator. Finally, it aggregates the information from the describing sentence and the related text to form a question or a claim. MQA-QG can generate data from tables or a hybrid of tables and texts. But the main deficiency is that it cannot integrate the information from multiple rows using complex underlying logic, so the generated questions/claims are relatively simple.

\lizy{(3) UCTR is a basic version of UCTR-ST, lacking the self-training process while keeping other components the same, and it cannot leverage unlabeled real data.}

(4) UCTR $-w/o$ T2T is an ablation model of UCTR. It represents the UCTR framework without the Table-To-Text and Text-To-Table operators, so it cannot generate samples containing both tabular and textual information as evidence.

(5) TAPAS-Transfer is a transfer learning model from TABFACT \cite{chen2019tabfact}. TABFACT is a large dataset focusing on fact verification on Wikipedia tables. It consists of 117,854 human-annotated claims on 16,573 tables. This model is trained on TABFACT and then directly applied on SEM-TAB-FACTS.


\begin{table*}[ht]
  \centering
  \caption{Results on the development set of TAT-QA}\scalebox{0.96}{
  \begin{tabular}{c|l|rr|rr|rr|rr}
  \toprule
 \multicolumn{2}{c|}{\multirow{2}{*}{Model}} & \multicolumn{2}{c|}{Table} &\multicolumn{2}{c|}{Table-Text}&\multicolumn{2}{c|}{Text}&\multicolumn{2}{c}{Total}\\\cmidrule{3-10}
  \multicolumn{2}{c|}{} & EM & F1&EM & F1&EM & F1&EM & F1  \\
\midrule
\multirow{4}{*}{ Supervised}
&Text-Span only  & 1.3&1.6& 7.7&9.7&47.3&73.5&14.0&20.9\\
& Table-Cell only   & 12.0 &16.8& 20.5&29.2&0.3&1.0&11.9&16.9\\
&TAPAS \cite{herzig2020tapas} &-&-&-&-&-&-&18.9&26.5 \\
&TAGOP  \cite{zhu2021tat} & \textbf{52.6} &\textbf{54.9}& \textbf{65.1}&\textbf{66.9} &\textbf{48.8}&\textbf{73.8}&\textbf{55.5}&\textbf{62.9}\\ \midrule 
\multirow{3}{*}{Unsupervised}
& MQA-QG \cite{pan2020unsupervised}&9.7&12.4&23.7 &30.1&33.2&55.1&19.4 & 27.7\\
& UCTR $-w/o$ T2T  & 28.1&30.0 &41.8& 47.1 & 30.6 &52.9 & 32.8 &40.5  \\
& UCTR (ours)&30.7&32.4 &42.8&47.3 & \textbf{33.2}&\textbf{55.9} & 34.9&42.4\\
& \lizy{\textbf{UCTR-ST (ours)}}&\textbf{38.2}&\textbf{40.3} &\textbf{50.3}&\textbf{54.7} & 31.1&52.8 & \textbf{40.2}&\textbf{47.6}\\
\midrule
\multirow{2}{*}{Few-Shot}
& TAGOP \cite{zhu2021tat}&10.4&13.4&11.2&18.6&0.3&0.9&8.3&12.1\\
&TAGOP+\textbf{UCTR-ST}&\textbf{42.9}&\textbf{45.7}&\textbf{58.8}&\textbf{62.4}&\textbf{44.5}&\textbf{72.1}&\textbf{48.1}&\textbf{56.9} \\
  \bottomrule
  \end{tabular}
  }
  \label{unsuper on tat}
\end{table*}

\begin{table*}[ht]
  \centering
  \caption{Results on Feverous}\scalebox{0.92}{
  \begin{tabular}{c|l|cc|c}
  \toprule
 \multicolumn{2}{c|}{\multirow{2}{*}{Model}} & \multicolumn{2}{c|}{Dev} &Test\\\cmidrule{3-5}
  \multicolumn{2}{c|}{} & Accuracy
  & FEVEROUS Score& FEVEROUS Score  \\
\midrule
\multirow{3}{*}{Supervised}&Sentence-only baseline &81.1&19.0&18.5\\
&Table-only baseline &81.6&19.1&17.9\\ 
&Full baseline \cite{aly2021feverous} & \textbf{86.0} &\textbf{20.2}&\textbf{19.2}\\ \midrule 
\multirow{3}{*}{Unsupervised} & Random &47.0&14.1& 13.2\\
& MQA-QG \cite{pan2020unsupervised}  &   71.1 & 17.6 &16.4 \\
& UCTR (ours) &74.8  &18.3& 17.0 \\
&\lizy{\textbf{UCTR-ST (ours)}}&\textbf{77.7}  &\textbf{19.7}& \textbf{18.3} \\
\midrule
\multirow{2}{*}{Few-Shot}
& Full baseline \cite{aly2021feverous}&67.3&14.2&13.3\\
& Full baseline+\textbf{UCTR-ST} &\textbf{78.2}&\textbf{19.7} &\textbf{18.4} \\
  \bottomrule
  
  \end{tabular}
  }
  \label{unsuper on feverous}
\end{table*}

\begin{table}[ht]
  \centering
  \caption{Results on SEM-TAB-FACTS }\scalebox{0.92}{
  \begin{tabular}{l|l|cr}
  \toprule
 \multicolumn{2}{c|}{\multirow{2}{*}{Model}} & \multicolumn{2}{l}{3-way micro F1}\\ 
  \multicolumn{2}{c|}{} &Dev&Test\\
\midrule
Supervised &TAPAS \cite{gautam2021volta} & \textbf{66.7} &\textbf{62.4} \\ \midrule
\multirow{4}{*}{Unsupervised} &Random &33.3&33.3\\
& MQA-QG \cite{pan2020unsupervised}   &53.2& 50.4\\
&TAPAS-Tranfer \cite{gautam2021volta}&59.0&58.7\\
& UCTR (ours)& 62.6 & 60.3 \\
&\lizy{\textbf{UCTR-ST (ours)}}& \textbf{64.2} & \textbf{61.2} \\
\midrule
\multirow{2}{*}{Few-Shot}
& TAPAS \cite{gautam2021volta} &48.6&46.5\\
& TAPAS+\textbf{UCTR-ST}&\textbf{64.1}&\textbf{61.0}  \\
  \bottomrule
  \end{tabular}
  }
  \label{unsuper on semeval}
\end{table}

\begin{table}[ht]
  \centering
  \caption{Results on WiKiSQL}\scalebox{0.92}{
  \begin{tabular}{l|l|cr}
  \toprule
 \multicolumn{2}{c|}{\multirow{2}{*}{Model}} & \multicolumn{2}{l}{Denotation Accuracy}\\ 
  \multicolumn{2}{c|}{} &Dev&Test\\
\midrule
\multirow{2}{*}{Supervised}&TAPAS \cite{herzig2020tapas} &85.1 &83.6 \\
&TAPEX \cite{liu2021tapex} & \textbf{88.1} &\textbf{87.0} \\ \midrule
\multirow{3}{*}{Unsupervised} &TAPEX \cite{liu2021tapex}  &21.4&21.8\\
& MQA-QG \cite{pan2020unsupervised}   &57.8& 57.2\\
& UCTR (ours)& 62.2& 61.6\\
& \lizy{\textbf{UCTR-ST (ours)}}& \textbf{63.5} & \textbf{62.7} \\
\midrule
\multirow{2}{*}{Few-Shot}
& TAPEX \cite{liu2021tapex} &53.8&52.9\\
& TAPEX+\textbf{UCTR-ST}&\textbf{63.5}&\textbf{62.7}  \\
  \bottomrule
  \end{tabular}
  }
  \label{unsuper on semeval}
\end{table}

 According to the results shown in Table \ref{unsuper on tat}, \ref{unsuper on feverous}, and \ref{unsuper on semeval}, we have the following observations:

(1) The basic version of our proposed framework--UCTR can already achieve promising unsupervised performance on the three datasets. Compared to supervised benchmarks, it reaches 67\%, 70\%, 87\%,  93\% of F1 score or label accuracy on the TAT-QA, WiKiSQL, FEVEROUS and SEM-TAB-FACTS, respectively, without using any human-labeled data. Moreover, UCTR outperforms other unsupervised models by large margins. In particular, the F1 score of MQA-QG on TAT-QA is only 27.7, while UCTR achieves 42.4. We suppose the reason is that the data generated by MQA-QG can only cover a small fraction of reasoning types compared to the original dataset, so the trained model cannot handle questions with more complex logical structures. Contrastively, UCTR can take advantage of program templates with various underlying reasoning structures to match the distribution of the original dataset as much as possible.

\lizy{
(2) Our proposed UCTR-ST framework achieves the best performance on all tasks. Compared to the basic version--UCTR, the self-training technique brings significant improvements, which suggests that although synthetic data is abundant, it differs from the distribution of realistic data. After incorporating the distribution information of realistic data by using unlabeled training samples, the model can perform better at testing stage.
}

(3) Under the few-shot setting, where only 50 labeled instances are available, supervised models perform poorly. In contrast, UCTR-ST gains much better performance with the assistance of a large amount of synthetic data. The results reveal that our method can significantly reduce the labor cost of manual annotation. Additionally, we notice that for FEVEROUS and TAT-QA, models trained on the synthetic dataset can gain further improvements by fine-tuning on the 50 high-quality sample
s. But for SEM-TAB-FACTS and WiKiSQL, the 50 human-labeled samples don't enhance the model as expected. We suppose it is because the 
  amount of annotated samples is too small to provide additional valuable information on these datasets.

(4) TAPAS-Transfer performs well without fine-tuning on any synthetic samples generated from SEM-TAB-FACTS, which reveals that sufficient training data from the general domain (i.e., the TABFACT dataset) can give a good model initialization for specialized domains. However, TAPAS-Transfer still underperforms our unsupervised framework UCTR-ST. We suppose there are two main reasons. Firstly, the samples of SEM-TAB-FACTS contain lots of scientific terms and numbers. In addition, SEM-TAB-FACTS has one more label--``Unknown'' compared to TABFACT, limiting the effectiveness of transfer learning from TABFACT.

\begin{table*}[ht]
  \centering
  \caption{Results of data augmentation on TAT-QA, FEVEROUS and SEM-TAB-FACTS }\scalebox{1}{
  \begin{tabular}{c|c|cc|cc|cc|c}
  \toprule
 \multicolumn{2}{c|}{\multirow{2}{*}{Model}} & \multicolumn{2}{c|}{TAT-QA} &\multicolumn{2}{c|}{SEM-TAB-FACTS}&\multicolumn{2}{c|}{WiKiSQL}&FEVEROUS\\
  \multicolumn{2}{c|}{} &Dev&Test&Dev&Test&Dev&Test &Dev \\
\midrule
\multirow{2}{*}{Supervised} &Baseline&55.5/62.9 &50.1/58.0&66.7&62.4&\textbf{88.1}&\textbf{87.0} &\textbf{86.0}\\ \cmidrule{2-9}
& Baseline+UCTR   &\textbf{59.7/67.7}&\textbf{56.1/64.3}&\textbf{69.8}&\textbf{63.9} &87.9&\textbf{87.0}&85.9\\
  \bottomrule
  \end{tabular}
  }
  \label{data aug}
\end{table*}

\subsection{Data Augmentation.}
In this section, we investigate the effectiveness of using our data generation method as a data augmentation technique. \lizy{Note that we assume that human-labeled training data is available, so we used the basic version--UCTR without using the self-training process.} We first fine-tune the model on our synthetic data and then fine-tune it on the high-quality human-labeled data. The performances are shown in Table \ref{data aug}.
For TAT-QA, the evaluation metric is the EM and F1 score. For WiKiSQL, the evaluation metric is the denotation accuracy. For FEVEROUS and SEM-TAB-FACTS, the evaluation metric is the label accuracy.
Experimental results show that the effectiveness of UCTR varies across different benchmarks. We surprisingly find that UCTR can substantially boost the supervised performance, with a 6.3 absolute gain of F1 score on the test set of TAT-QA and 3.1 gain of label accuracy on the development set of SEM-TAB-FACTS. But similar phenomena are not observed for the FEVEROUS and WiKiSQL. 

We suppose the main underlying reason is that UCTR can alleviate the problem of data sparsity. Both TAT-QA and SEM-TAB-FACTS are datasets collected from specialized domains. And the labeled training samples of them are relatively insufficient. Specifically, the number of tables in TAT-QA and SEM-TAB-FACTS are 2,757, 1,085, respectively, compared to over ten thousand tables for FEVEROUS and WiKiSQL. As a result, the data generated by UCTR can make the model get familiar with the tables and provide a good initialization for supervised training.

In summary, our proposed unified framework can generate high-quality human-like data for various tabular reasoning tasks on homogenous or heterogeneous data. The synthetic data can significantly boost the model's performance under an unsupervised or a few-shot setting and even enhance the supervised performance further.

\begin{table*}[ht]
  \centering
  \caption{Ablations on the development set of TAT-QA }\scalebox{0.92}{
  \begin{tabular}{c|c|c|c|c|c|c|c|c|c|c}
  \toprule
  
\multirow{3}{*}{Setting}&\multicolumn{3}{c|}{Data Source}&\multicolumn{2}{c|}{Program Type}& Self & \multicolumn{4}{c}{Performance}\\\cline{2-6}\cline{8-11}
 
&\multirow{2}{*}{Table}& \multirow{2}{*}{Text} &\multirow{2}{*}{Table$\leftrightarrow $Text} &\multirow{2}{*}{SQL}&\multirow{2}{*}{Arithmetic}&\multirow{2}{*}{Training}&Table&Table-Text&Text&Total\\ \cline{8-11}
   
 & &&&&   &&$EM$\ /\ $F_1$&$EM$\ /\ $F_1$& $EM$\ /\ $F_1$& $EM$\ /\ $F_1$\\\midrule
 
  A1&\checkmark& &&\checkmark & &&6.1\ /\ 8.6 &17.2\ /\ 21.6& 0.8 /\ 1.5&8.2\ /\ 10.9 \\
 
 A2& &\checkmark&&&& &1.8\ /\ 2.2 &5.3\ /\ 8.2& 32.1 /\ 55.8&10.0\ /\ 16.5\\
 A3&\checkmark &\checkmark&&\checkmark& &&6.3\ /\ 8.4 &17.8\ /\ 23.4 & 31.4 /\ 54.1&15.7\ /\ 23.6\\
 A4& \checkmark &\checkmark&&&\checkmark& &30.6\ /\ 31.7 &35.9\ /\ 38.8 & 31.8 /\ 53.0&32.5\ /\ 38.8\\
 A5&\checkmark &\checkmark&&\checkmark&\checkmark& &28.1\ /\ 30.0 &41.8\ /\ 47.1 & 30.6 /\ 52.9 & 32.8 /\ 40.5\\
 A6& \checkmark &\checkmark&\checkmark&\checkmark&\checkmark& &30.7\ /\ 32.4 &42.8\ /\ 47.3 & 33.2 /\ 55.9 & 34.9 /\ 42.4\\
A7& \checkmark &\checkmark&\checkmark&\checkmark&\checkmark&\checkmark &38.2\ /\ 40.3 &50.3\ /\ 54.7 & 31.1 /\ 52.8 & 40.2 /\ 47.6\\
  \bottomrule
  \end{tabular} 
  }
  \label{ablation}
\end{table*}

\begin{table*}[h]
    \centering
    \caption{Examples exhibiting generated text from different types of programs. The red spans are the key information shared by the generated text and golden text, while the spans colored by blue are the information mismatched.}
\scalebox{0.89}{
    \begin{tabular}{m{0.4\columnwidth}m{0.6\columnwidth}m{0.5\columnwidth}<{\centering}m{0.5\columnwidth}<{\centering}}
    \toprule
        \textbf{Type} & \textbf{Program}  & \textbf{Generated Text} &\textbf{Golden Text}\\
    \midrule
    
SQL Query & select [department] from table order by [total deputies] desc limit 1
& Which \textcolor{red}{department} has the \textcolor{red}{most total deputies}? \ & What is the \textcolor{red}{department} with the \textcolor{red}{most} amount of \textcolor{red}{total deputies}? \\ \midrule

 Logical Form  & eq \{ count \{ filter\_all \{ all\_rows ; Basic Printer Settings
Material \} \} ; 3 \}
& There are \textcolor{red}{3 basic printer settings} that can be used with a Basic Printer \ & 
There are \textcolor{red}{3} \textcolor{blue}{Material} used for \textcolor{red}{Basic Printer Settings}.   \\ \midrule

Arithmetic Expression & subtract ( the Stockholders' equity of 2019, the Stockholders' equity of 2018 ) , divide ( \#0, the Stockholders' equity of 2018 )
& By what \textcolor{red}{percentage} did stockholders' equity \textcolor{blue}{decrease} from \textcolor{red}{2018} to \textcolor{red}{2019}? & What was the \textcolor{red}{percentage} change in stockholders' equity between \textcolor{red}{2018} and \textcolor{red}{2019}?\\ \midrule

    \end{tabular}
    }
    \label{tab:my_label}
\end{table*}

\subsection{Ablation Study.}
To evaluate the effectiveness of each component of UCTR-ST, we present the model's performances on the development set of TAT-QA under different ablation settings. The results are depicted in Table \ref{ablation}. ``Table$\leftrightarrow$ Text" under the ``Data Source" column means we generate joint table-text reasoning samples using the Table-To-Text operator and Text-To-Table operator. Based on the results, we have the following observations:

From the perspective of data sources, models trained only on tables or texts achieve low performances. In contrast, the model trained on both tabular and textual data obtains the capability of reasoning across modalities and gains considerable improvement. Additionally, the ``Table$\leftrightarrow$ Text" source brings further enhancement, highlighting the ability to reason on a hybrid of tabular and textual data.

From the perspective of program types, arithmetic expressions are more valuable than SQL queries  since most samples in TAT-QA require arithmetic operations. The model using all these two types of programs reaches the highest performance.
\subsection{Analysis of Generated Text.}
In this section, we present some example sentences generated by the Program-Transformation from different types of programs. The red text spans are important information shared by generated and golden text, while the blue spans are the information mismatched. We can observe that Program-Transformation module can understand the underlying logic and generate appropriate questions or claims. For example, the original arithmetic expression only contains a ``subtract" operation followed by a ``divide" operation, but the model can identify the meaning of ``percentage change" correctly. However, in some cases, the generated text loses some critical information or contains inaccurate information.

\subsection{Synthetic Data vs. Labeled Data.}

\begin{figure}[hb]
  \centering
\includegraphics[width=0.8\linewidth]{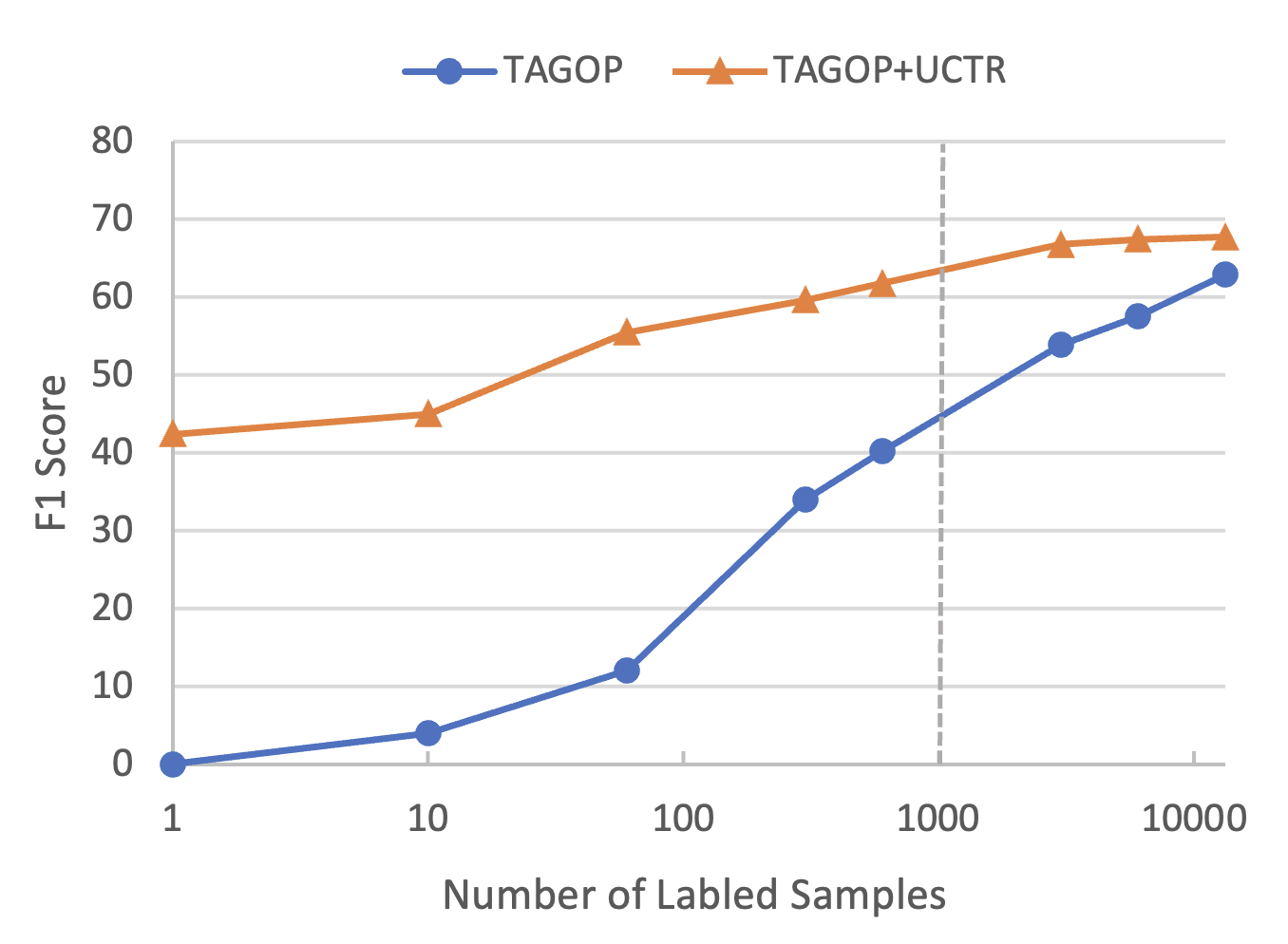}
  \caption{Effectiveness of the synthetic data. The orange line corresponds to the model first trained on the synthetic data and then fine-tuned on the varied number of labeled samples. The blue line corresponds to the model directly trained on labeled samples. }
  \label{few-shot-tat}
\end{figure}

In section \ref{results}, we show the few-shot performance of models using only 50 samples. In this section, we conduct a more detailed analysis of the synthetic data and labeled data by changing the number of available labeled samples. Since the synthetic data shows significant effects on TAT-QA according to previous results, we still take the result on the development set of TAT-QA as an example. As shown in Figure \ref{few-shot-tat}, the orange line depicts the F1 score of the model first trained on our synthetic data, then further fine-tuned on the available labeled data. In contrast, the blue line shows the performance of the model directly trained on the labeled data.

As the number of samples increases,  the model pre-trained on our synthetic data always performs better. In addition, we have several interesting findings: i) The F1 score of the model trained on 23,933 synthetic samples is around 42, comparable to a model trained on 1000 labeled data. (2) When we fine-tune the model trained on 23,933 synthetic samples with additional 1000 human-labeled samples, it can achieve comparable performance to a model trained on 13,217 labeled data. Therefore, we conclude that our unsupervised learning framework provides a good initialization so that the model can gain a considerable improvement using only a small amount of labeled samples. Our framework can be very beneficial in an online learning setting when applying a model to a new domain, where labeled data is limited.

\begin{figure*}[h]
  \centering
  \includegraphics[width=0.7\linewidth]{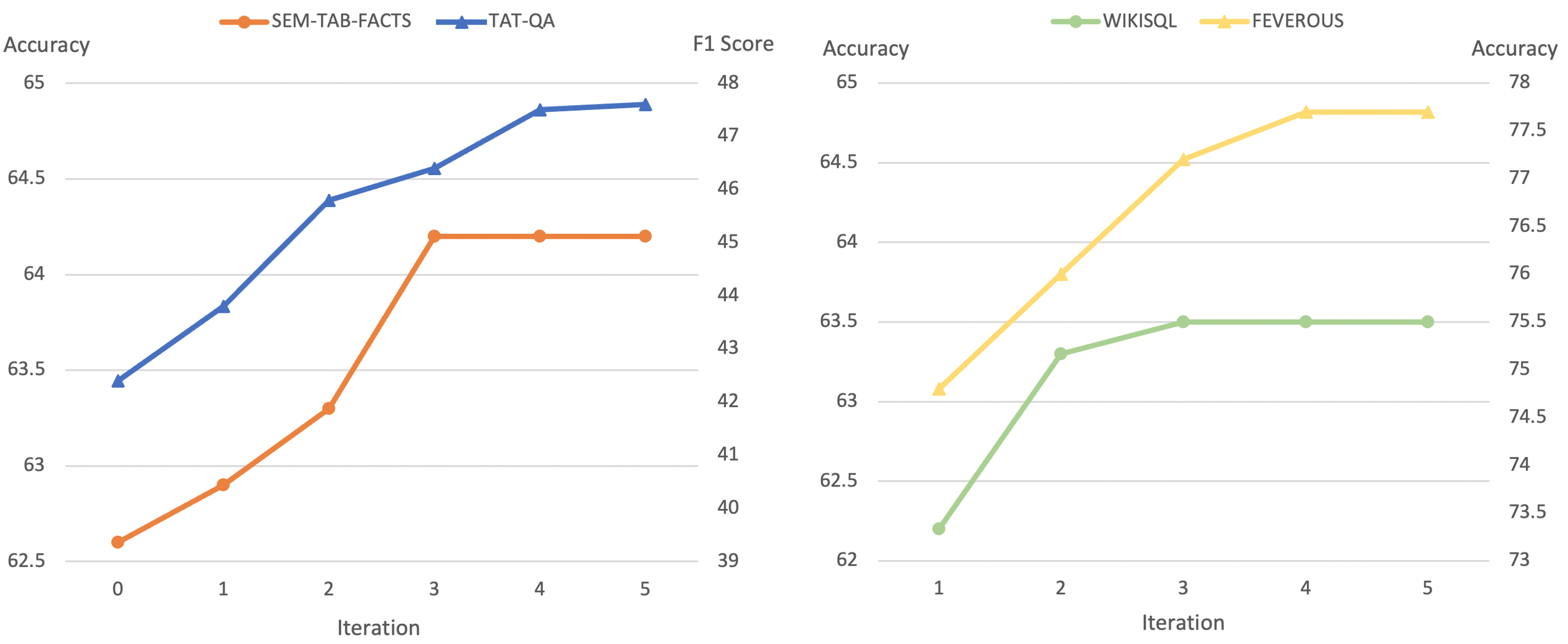}
  \caption{\lizy{Effectiveness of the self-training on four benchmarks. We demonstrate how the performance of the model varies with the number of iterations of self-training. It can be observed that the model can self-improve its performance by using only unlabeled data and eventually achieves convergence in about 6 iterations. The final model shows a significant improvement compared to the original model.}}
  \label{self-training-by-epoch}
\end{figure*}

\subsection{\lizy{Effectiveness of Self-Training}}

\lizy{
In order to better demonstrate the role of self-training technique in the training process, we show in Figure \ref{self-training-by-epoch} the performance of the models on four benchmarks as the number of iterations varies. It can be seen that as the model performance improves, the model can generate better pseudo-labels, which in turn can be used to train better models in the next iteration. As the number of iterations increases, the improvement of the model will gradually decrease, and eventually converge. In a nutshell, we verify that our data generation technique can naturally integrate with self-training in a fully unsupervised scenario and achieve good performance on real world data.}

\section{RELATED WORK}
In this section, we briefly summarize the related works from these two aspects: the development of tabular reasoning models and
 unsupervised data generation methods.  

\subsection{Tabular Reasoning Models.}
Many tabular reasoning models tackle the question answering and fact verification tasks in a semantic parsing manner \cite{guu2017language}, \cite{liang2016neural}, converting a natural language sentence into a program. Zhong et al. \cite{zhong2017seq2sql} translate users' questions to corresponding SQL queries, and Yang et al. \cite{yang2020program} generate semantic consistent logical forms with tree structures and execute them to judge the claims. However, the search space for programs is very large, and the model may generate spurious programs which have wrong structures but return the correct answers.
Recent works demonstrate that pre-trained language models achieve better reasoning performances on various tasks by pre-training or leveraging auxiliary knowledge   \cite{glass2021capturing}, \cite{duan2021bridging},  \cite{duan2022not}, \cite{li2021jointly}. Specifically, for the tabular reasoning task, TAPAS \cite{herzig2020tapas} is a BERT-extended model pre-trained on a large corpus of texts and tables from Wikipedia. It answers questions by applying operations on predicted table cells in an end-to-end way. Neeraja et al. \cite{neeraja2021incorporating} boost the reasoning ability of pre-trained models on the tabular NLI task by introducing external knowledge. And it is a promising direction to explore how to obtain better representations of tables. GraPPa \cite{yu2020grappa} introduces a text-schema linking objective to make the model better understand the grammatical role of table elements. However, the main drawback of these methods is that they require a large amount of training data, limiting their performance when transferring to a new domain.

\subsection{Unsupervised Data Generation and Self-Training.}
Unsupervised data generation has been extensively studied on various tasks like question answering and natural language inference, and has shown surprising performances \cite{varshney2021unsupervised}, \cite{dong2023toward} \cite{li2023toward}. Recently, methods for synthesizing human-like tabular reasoning samples have also been proposed \cite{guo2018question}, \cite{serban2016generating}. Chemmengath et al. \cite{chemmengath2021topic} sample complex SQL queries and generate natural language questions in a seq2seq manner. Eisenschlos et al. \cite{eisenschlos2020understanding} generate factual claims leveraging context-free grammar (CFG) and counterfactual heuristics. Unfortunately, these methods focus on a specific task or scenario. Based on the modules and predefined operators, our approach can convert different types of programs into natural language questions or claims with tabular evidence or a hybrid of tabular and textual evidence.

Self-training has been widely explored in the realm of semi-supervised learning \cite{li2022effective}, \cite{grandvalet2004semi}, \cite{lee2013pseudo}. For example, Li et al. \cite{li2024flexkbqa} proposed FlexKBQA, a method that combines self-training and synthetic data to improve the performance of few-shot knowledge based question answering. Most works employ self-training techniques in the few-shot setting, where a small number of labeled samples are available. However, this study effectively combines data generation methods with self-training to achieve good results in an unsupervised scenario.

\section{Conclusion and future works}\label{sec-conclusion}
We explore the unsupervised complex tabular reasoning task and propose a novel self-training framework UCTR-ST with several optimization techniques. UCTR-ST can synthesize high-quality human-like questions and claims with underlying complex logic without any labeled data and leverage the self-training technique to fully exploit the information of the unlabeled data. Comprehensive experiments for different tasks and domains demonstrate that model achieve surprising performances under unsupervised and few-shot settings, which can significantly ease the burden of human annotation. Moreover, UCTR-ST can boost supervised performances in specialized domains with insufficient data. In future work, we will broaden the reasoning types of programs and explore an auto program-generation method based on the existing data distributions to make the framework more flexible.

\section*{Acknowledgment}
This work was supported in part by National Key Research and Development Program of China under Grant No. 2020YFA0804503, National Natural Science Foundation of China under Grant No. 62272264, and Beijing Academy of Artificial Intelligence (BAAI).


 
\vspace{11pt}

\vspace{11pt}


\vfill

\end{document}